# Attempto

# From Specifications in Controlled Natural Language towards Executable Specifications

Rolf Schwitter, Norbert E. Fuchs

Department of Computer Science, University of Zurich
{schwitter, fuchs}@ifi.unizh.ch

Deriving formal specifications from informal requirements is difficult since one has to take into account the disparate conceptual worlds of the application domain and of software development. To bridge the conceptual gap we propose controlled natural language as a textual view on formal specifications in logic. The specification language Attempto Controlled English (ACE) is a subset of natural language that can be accurately and efficiently processed by a computer, but is expressive enough to allow natural usage. The Attempto system translates specifications in ACE into discourse representation structures and into Prolog. The resulting knowledge base can be queried in ACE for verification, and it can be executed for simulation, prototyping and validation of the specification.

# 1    Bridging the Gap between Informal and Formal Methods

The derivation of formal specifications from informal requirements is difficult, and known to be crucial for the subsequent software development process. Requirements emerge from heterogeneous sources and express different and even varying viewpoints of the domain specialists. Too often requirements are informal, vague, contradictory, incomplete, and ambiguous. To derive formal specifications from deficient requirements of that kind is not an easy task.

Nevertheless, we can help a lame dog over a stile. To bridge the conceptual gap between application domains and formal specifications we introduce graphical and textual views as application-oriented representations of formal specifications. Graphical and textual views allow domain specialists and software engineers to express their concepts of the application domain concisely and directly. By making "true statements about the intended domain of discourse" [Kramer & Mylopoulos 92] and "expressing basic concepts directly, without encoding, taking the objects of the language as abstract entities" [Börger & Rosenzweig 94] domain specialists can immediately convince themselves of the adequacy of the specification.

An automatic mapping between a view and its associated formal specification in a logic language assigns a formal semantics to the view. Though views give the impression of being informal and having no intrinsic meaning, they are formal and have the semantics of their associated formal specification. This dual-faced appearance of views reduces the conceptual gap and allow users to compose and to examine specifications in concepts of the application domain.

Since the formal specification is executable the execution can be observed on the level of the view, and validation and prototyping in concepts close to the application domain become possible.

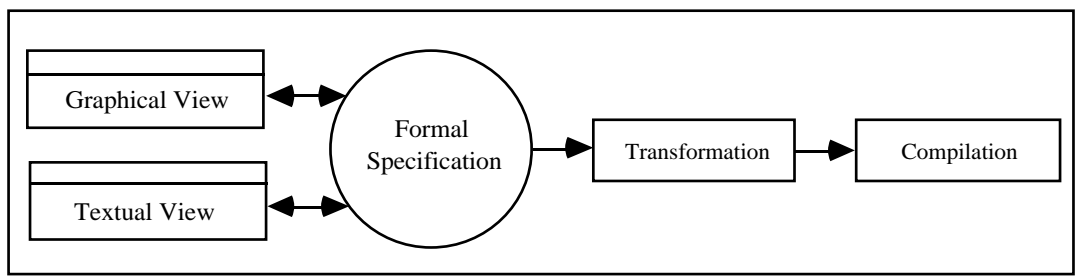

Though it is executable the formal specification is not necessarily an efficient program. To derive an efficient program the specification is transformed by powerful source-to-source transformations before it is compiled. A schema-based transformation system for logic programs is described in [Vasconcelos & Fuchs 95].

Graphical views that include transition networks for finite state automata and window oriented user-interfaces are presented in [Fuchs & Fromherz 94].

In this paper we introduce controlled natural language – concretely Attempto Controlled English (ACE) – as a textual view for writing functional requirements specifications. First of all, we motivate the use of controlled natural language and present ACE (section 2); then, we give an overview of the specification system Attempto (section 3); after describing the parsing process (section 4), we discuss the translation from a specification in ACE to Prolog via Discourse Representation Theory (section 5); given this translation, we investigate some issues concerning the execution of the specification (section 6). Finally, we conclude and outline further research (section 7).

## 2   Controlled Natural Language as a Textual View

### 2.1   Why Controlled Natural Language?

Natural language as a familiar means of communication has a long tradition in requirements engineering. However, uncontrolled use of natural language leads to ambiguous, imprecise and unclear specifications. Beyond that, requirements vary and new requirements arise so that the specification is subject to frequent change. People have advocated the use of formal specification languages to eliminate some of the problems associated with natural language. However, because of the need of comprehensibility, we cannot replace documents written in natural language by formal specifications in all cases. Many domain specialists would not understand such a formal document and would hardly accept it as a contract for the software system [Sommerville 92].

Though it seems that we are stuck between the over-flexibility of natural language and the potential incomprehensibility of formal languages, there is a way out. To improve the quality of specifications without losing their readability, we propose to restrict natural language to a controlled subset with a well-defined syntax and semantics that can serve as a suitable view of a logic language. On the one hand this subset should be expressive enough to allow natural usage by domain specialists, and on the other hand the language should be accurately and efficiently processable by a computer. This means that we have to find the right trade-off between expressiveness and processability [Allen 91, Pulman 94].

Controlled or simplified English is not a new idea. It has been used for quite some time for technical documentation [Wojcik et al. 90, Adriaens & Schreurs 92, AECMA 95], and lately both for knowledge-based machine translation in the KANT system [Mitamura & Nyberg 95] and as data base query language [Androutsopoulos 95]. Additionally, a general computer processable controlled language has been suggested that could be used for various purposes ranging from structured documentation over access to information to the control of devices [Pulman & Rayner 94]. However, very few researchers have tried to employ controlled natural language for software specifications since this leads to further syntactic and semantic constraints for the language, especially if one requires the specifications to be executable [Ishihara et al. 92, Macias & Pulman 92, Pulman 94, Fuchs & Schwitter 95].

Domain specialists seem to be able to construct sentences in controlled natural language, and to avoid constructs that fall outside the bounds of the language, particularly when the system gives feedback of the analysed sentences in a paraphrased form using the same controlled language [Capindale & Crawford 89].

## 2.2 Attempto Controlled English (ACE)

As a starting-point we suggest the following basic model of ACE [Fuchs & Schwitter 96]. In ACE declarative sentences can be combined by constructors to powerful composite sentences, while certain forms of anaphora and ellipsis leave the language concise and natural. Furthermore, we place interrogative sentences at the user's disposal for verifying the translated specification text.

Specification texts consist of

- declarative sentences *subject + finite verb (+ complement or object)*
- composite sentences built from simpler sentences with the help of constructors that mark coordination (*and*, *or*, *either-or*), subordination (*if-then*, *who/which/that*), negation (*not*), and negated coordination (*neither-nor*)

Sentences can contain

- subject and object modifying relative sentences
- anaphoric references, e.g. *personal pronouns*
- coordination between equal constituents, e.g. *and, or*
- ellipsis as reduction of coordination
- negated noun phrases, *no X*
- synonyms and abbreviations

Interrogative sentences comprise

- *yes/no*-questions
- *wh*-questions

In ACE finite verbs can only be used actively, in the simple present tense, and in their third person singular and plural forms. In this way, users are supported to express statements that are always true and refer to events or states which are true in the present period of time, or possibly true when used in conditional sentences.

## 2.3 Example Specification in ACE

The following example is a small excerpt of an ACE specification for a simple automated teller machine called SimpleMat.

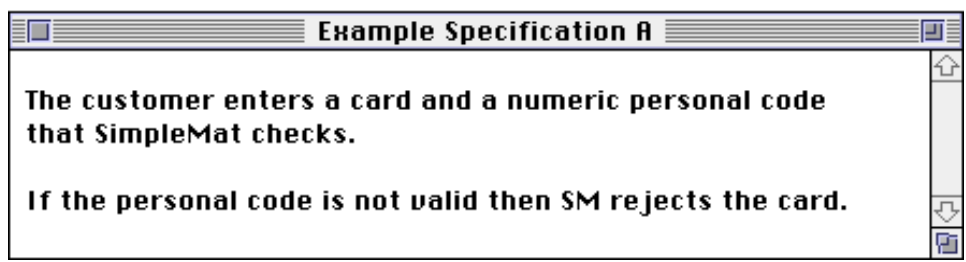

The specification text employs

- composite sentences built from declarative sentences with the help of the coordinator *and*, the subordinators *that* and *if-then*, and the negation *not*
- compound nouns, e.g. `personal code`
- ellipsis
- anaphoric reference by definite noun phrase, e.g. `the personal code`
- abbreviation (`SM` standing for SimpleMat)

## 3 Overview of Attempto

### 3.1 Translation Components

In this section we briefly describe the components of the specification system Attempto that we have implemented. Attempto accepts specifications in ACE and translates them into discourse representation structures (a structured form of first-order predicate logic), and then into Prolog.

The user enters specification text in ACE that the *Dialog Component* forwards to the parser in tokenised form. Parsing errors and ambiguities to be resolved by the user are reported back by the dialog component. The *Parser* uses a predefined Definite Clause Grammar enhanced by feature structures and a predefined linguistic lexicon to check sentences for syntactical correctness, and to generate syntax trees and sets of nested discourse representation structures. The *Linguistic Lexicon* contains an application-specific vocabulary. The lexicon can be modified by a lexical editor invokable from the dialog component – if necessary during the parsing process. The *Discourse Handler* analyses and resolves inter-text references and updates the discourse representation structures generated by the parser. The *Translator* translates discourse representation structures into Prolog clauses. These Prolog clauses are either passed to the knowledge assimilator, or – in case of queries – to the inference engine. The *Knowledge Assimilator* adds new knowledge incrementally to the knowledge base. The user can submit queries in ACE to the dialog component which are processed similarly to the specification text.

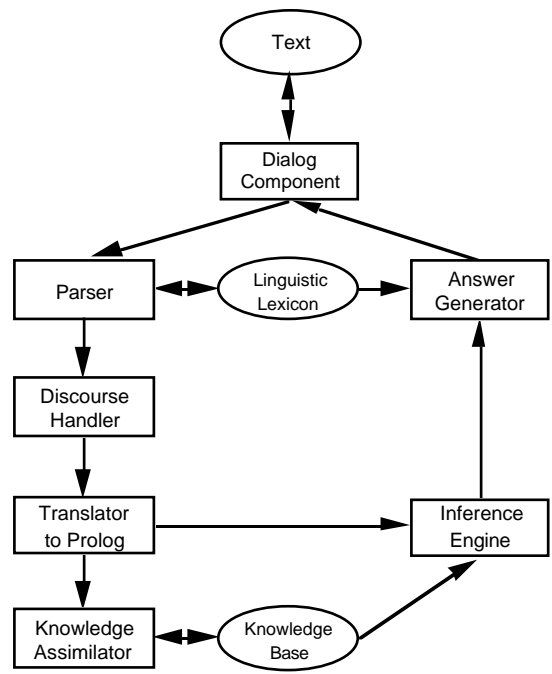

The *Inference Engine* answers user queries with the help of the knowledge base. In a preliminary version the inference engine is just the Prolog interpreter. The *Answer Generator* takes the answers of the inference engine, reformulates them in ACE, and forwards them to the dialog component.

## 3.2  Tools: Lexical Editor and Spelling Checker

Specification texts are gradually developed by domain specialists. Though Attempto's lexicon contains entries of the function word class, e.g. determiners, prepositions, pronouns and conjunctions, the entries for domain specific subsets of the content word class, e.g. nouns, verbs, adjectives and adverbs have to be added incrementally as needed for the specification text. A *Lexical Editor* – exhibiting interfaces for linguistic experts and non-experts – allows users to interactively modify and extend the full-form lexicon while the system parses the specification text.

The expert interface represents lexical entries as complete feature structures and allows experts to freely modify any lexical entry. The interface for non-experts employs templates that help users to enter information – presupposing minimal linguistic knowledge. The rest of the information is automatically derived. Help texts and balloon help support both groups of users.

The following screen shot shows how a non-expert would add the common noun `customer` to the lexicon. The user has to enter the singular and the plural forms, to select the gender and to decide between the classes of count nouns and mass nouns. Optionally, the user can connect each word with a number of synonyms or abbreviations.

```
Lexical entry: common noun
Please enter the common noun:
Form of the noun:
Singular:  customer
Plural:    customers

Gender of the noun:
 ○ feminine    ○ masculine    ● fem/masc    ○ neutrum

The nature of the noun:
 ● count noun    ○ mass noun

Help on: [ ... ▼ ]     [Synonyms]  [Cancel]  [Continue]
```

Attempto shifts the responsibility to check for lexical ambiguity onto the users; they decide how to use a word in a specification text – and bear the consequences.

Another tool is the *Spelling Checker* that allows users to determine whether all words of a specification text are in the lexicon. This spelling checker is invoked automatically if (part of) a specification text cannot be parsed.

## 4  Parsing

The specification text is parsed by a top-down recursive-descent parser coming free with Prolog. It uses a Definite Clause Grammar enhanced by feature structures written in GULP (Graph Unification Logic Programming) – a syntactic extension of Prolog that supports the implementation of unification-based phrase structure rules by adding a notation for linearised feature structures [Covington 94]. GULP adds to Prolog two operators and a number of system predicates. The first operator ':' binds a feature name to its value that can be another feature structure. The second operator '..' joins one feature-value pair to the next. Thus, we can write grammar rules like

```
sentence -->   noun_phrase(case:nom .. agr:Person_Number),
               verb_phrase(agr:Person_Number).
```

where the argument stands for a complex feature structure in which the feature `case` has the value `nom` for nominative and the feature `agr` enforces agreement of person and number between the noun phrase and the verb phrase.

The parser generates a syntax tree as syntactic representation, and concurrently a discourse representation structure as semantic representation.

In addition, the parser builds the following paraphrase for the example specification of section 2.3. Displaying all substitutions and interpretations made, the paraphrase reflects how Attempto interpreted the user's input.

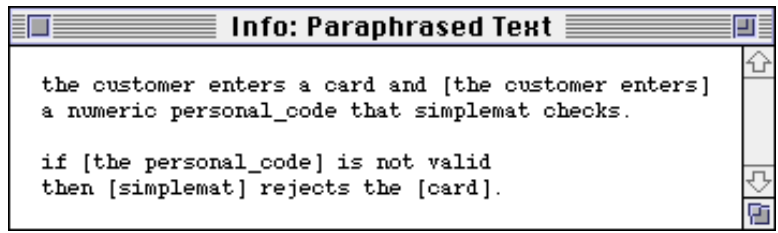

The user can now decide to accept Attempto's interpretation, or to rephrase the input to achieve another interpretation. For ambiguous input Attempto always suggest one interpretation as default. It is up to the user to reformulate the input to achieve another interpretation. If the user intends to express that the money dispenser should check both the card and the personal code he would have to formulate this intended meaning explicitly, for instance as

```
The customer enters a card and a numeric personal code.
SimpleMat checks the card and the personal code.
```

In addition, the parser informs the user about spelling and parsing errors. If the spelling checker finds an unknown word the user can immediately add the word to the lexicon with the help of the lexical editor and resubmit the input to the parser.

## 5   Semantic Representation via Discourse Representation Theory

Correct understanding of a specification requires not only processing of individual sentences and their constituents, but also taking into account the way sentences are interrelated to express complex propositional structures. We do this by employing Discourse Representation Theory (DRT) [Kamp & Reyle 93], and by extending our parser to extract the semantic structure of a sentence in the context of the preceding sentences [Covington et al. 88].

DRT represents a multisentential natural language discourse in a single logical unit called a discourse representation structure (DRS). The specification text is translated into a DRS which contains discourse referents representing the objects of the discourse, and conditions for these discourse referents. A DRS gives some kind of schematic picture of the real world that is true if the discourse referents can be taken as representations of real objects satisfying the conditions.

The picture shows that the first sentence of our example specification (section 2.3) contributes the discourse referents [A, B, C, D] and a number of conditions – with white background – to the DRS.

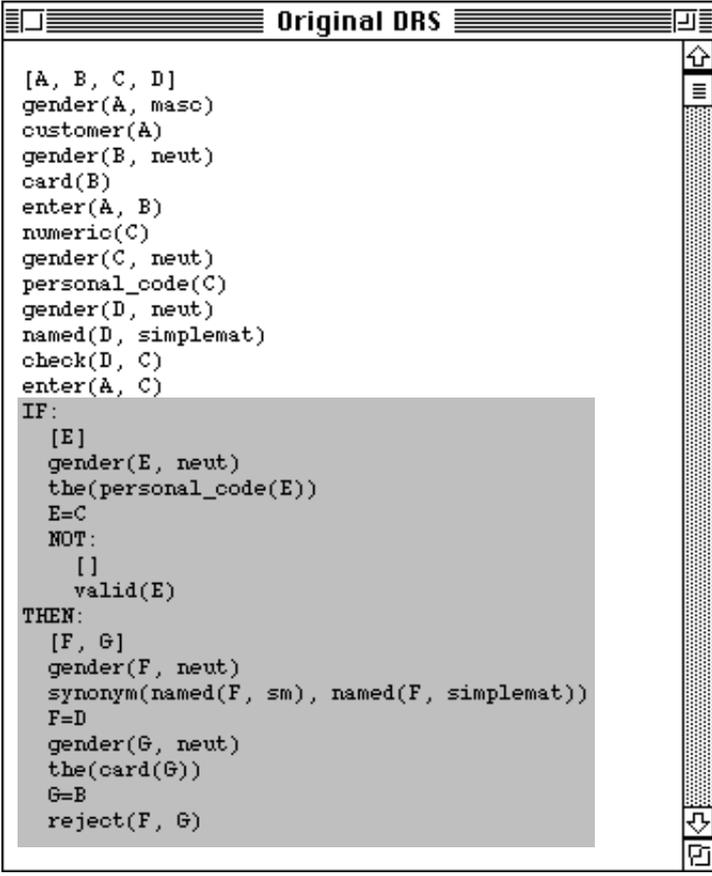

The second sentence is analysed in the context of the first sentence thus making the resolution of anaphoric references possible. This sentence contributes further discourse referents E, F, G, and the conditions with shaded background.

Conditions can be simple – e.g. `customer(A)` – or complex, i.e. again a DRS. This can lead to nested DRSs. In our case, the topmost DRS for the second sentence contains an `IF-THEN` sub-DRS which itself contains a `NOT` sub-DRS.

Note that discourse referents and conditions for proper names, e.g. `named(D, simplemat)` appear always in the topmost DRS.

Anaphoric references, e.g. the definite noun phrase `the personal code` of the second sentence referring to the phrase already introduced in the first sentence as `the numeric personal code`, are represented as conjunctive conditions `the(personal_code(E))` and `E=C`. Synonyms are treated similarly, they are associated with their reference word in the superordinated DRS, e.g. `synonym(named(F,sm),named(F,simplemat))` and `F=D`.

References are only possible to discourse referents in superordinated DRSs. The resolution algorithm always picks the closest referent that agrees in gender and number (the number feature is not displayed in the DRS).

In a further step the DRS is cleaned up – gender information that was only necessary for anaphoric resolution is eliminated, and all unifications are performed – yielding the final semantic representation of the complete specification text as one (nested) DRS

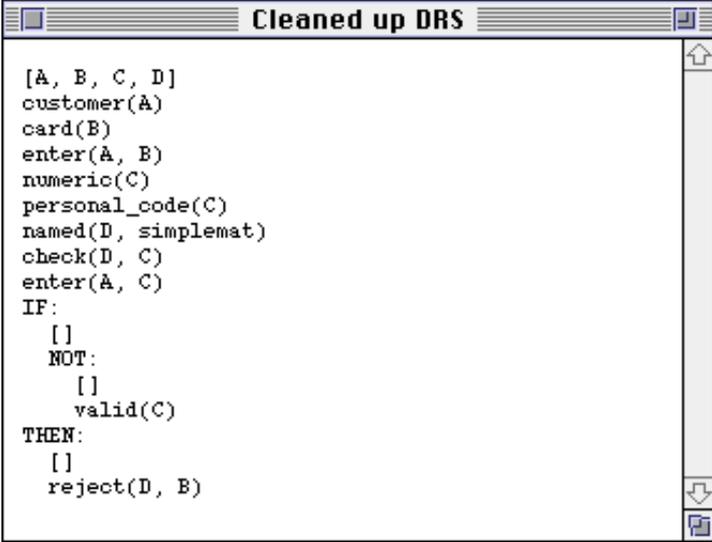

Finally, the discourse representation structure is translated into Prolog clauses which are asserted as `fact/1` to the knowledge base.

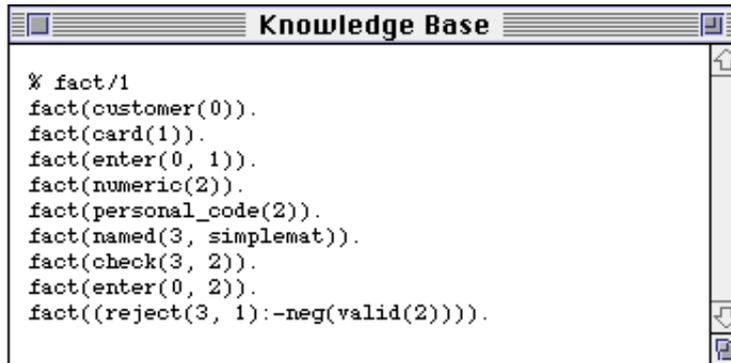

Discourse referents – being existentially quantified variables – are replaced by Skolem constants `0, 1, ...` or – if they are in the scope of a universal quantor – by Skolem functions.

`IF-THEN` DRSs with disjunctive consequences cannot directly be translated into Prolog since they would lead to disjunctive clauses. Instead they are represented by sets of Prolog clauses, one clause for each disjunct.

Interrogative sentences (*yes/no*-questions and *wh*-questions) can be used to query the contents of the knowledge base. Questions are translated first into `QUERY` DRSs and then into Prolog queries, and are answered by logical inference.

## 6   Towards Executable Specifications

### 6.1   Order of Events, Side-Effects, and Situation Specific Information

The knowledge base can be used for simulation or prototyping by executing it. In our example specification, this means executing the specification of the automated teller machine. As it stands, however, the specification does not provide all the necessary information and needs to be enhanced in three ways.

First and most importantly, a chronological order of events has to be established, e.g. we have to make sure that during the simulation the event of entering a card has to precede the event of checking the personal code. [Ishihara et al. 92] who translate natural language specifications into algebraic ones use contextual dependency and properties of data types to establish the correct order of events. In our approach based on Discourse Representation Theory the order of events is to a great extent established when we introduce eventualities (events and states) into the processing of our controlled natural language. Following Kamp [Kamp & Reyle 93] we interpret the text

```
The customer enters a card and a personal code
that SimpleMat checks.
```

as an *entering* event temporally followed by a *checking* event. Thus the sequence of the sentences of the specification text leads automatically to an ordering of events. On the basis of this information a simple meta-interpreter can execute the conditions of the DRS generated for the specification text in their correct order.

Second, many relations representing events, e.g. I/O operations, are not only truth-functional, but also cause side-effects. These side-effects can be defined by interface predicates in the form of Prolog clauses, e.g. for the verb `checks`

```
check(X, Y) :- write(['event: ', X, 'checks', Y]).
```

that depend on the simulation environment. The above interface predicate just traces events, one could, however, envisage that the interface predicates do not simply simulate the automated teller machine but cause the execution of a real automated teller machine.

Finally, the execution needs some situation specific information, or scaffolding. For example, to execute the specification

```
Every customer has a card.
```

the condition `customer(X)` must be provable for an instance of `X`. We can either provide this information as a statement in an ACE definition file, e.g.

```
John is a customer.
```

or more conveniently, get the information by querying the user.

### 6.2  Executing the Specification

If the user enters the specification text its execution needs situation specific information and queries the user, e.g. to get an instance of a specific customer.

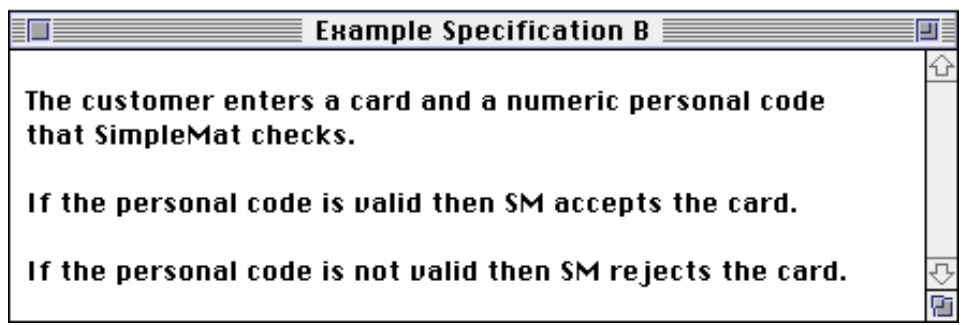

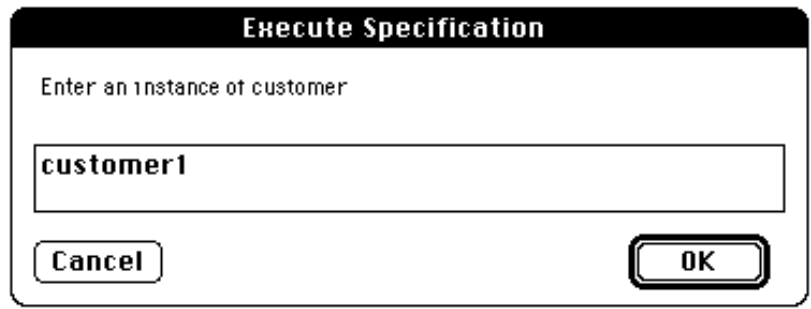

The execution of the above specification text leads to the following dialog with the user. Side-effects of events are simulated by simply printing out a trace, informing the user – step-by-step – about the relevant events that have been triggered.

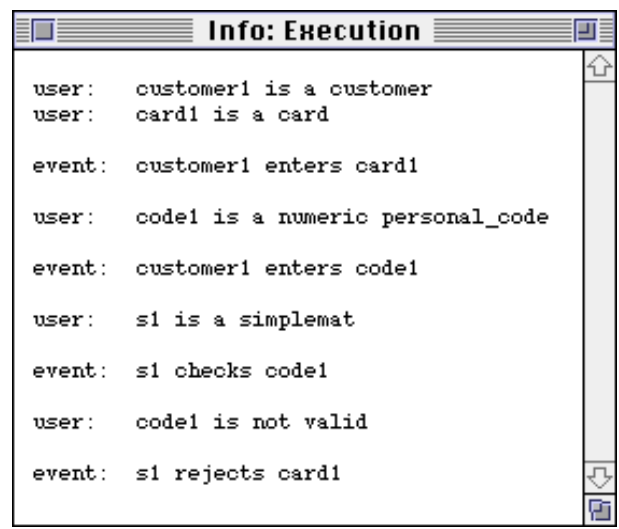

## 7    Conclusion and Future Research

The present prototypical implementation of our specification system Attempto proves that ACE can be used for the non-trivial specification of an automated teller machine. The specification can be translated as coherent text into Prolog via Discourse Representation Theory, and it can be executed.

More work needs to be done, however, to turn the prototype into a useful tool. Our current version of ACE was derived experimentally to represent typical constructs in natural language specifications in a structured and concise way. But, what we need, is a more systematic definition of ACE based on a small number of easily remembered principles. Furthermore, we plan to optimise the parsing strategy and to extend the internal syntactic and semantic representation. Especially, we intend to introduce event semantics into the DRSs to express the order of events in a more fine-grained manner, and expect an advantageous effect on the execution of the specification text.

Though natural language – even in a controlled form – is a universal specification notation, we believe that it is not always the optimal one. Thus we will bring together controlled natural language and graphical notations, e.g. for window-oriented user interfaces, and specific notations for algorithms.

# References

[Adriaens & Schreurs 92]　G. Adriaens, D. Schreurs, From Cogram to Alcogram: Towards a Controlled English Grammar Checker, Proceedings COLING 92, pp. 595-601, 1992

[Allen 91]　J. F. Allen, Natural Language, Knowledge Representation, and Logical Form, Technical Report 367, Computer Science Department, University of Rochester, 1991

[AECMA 95]　The European Association of Aerospace Industries (AECMA), AECMA Simplified English, AECMA Document: PSC-85-16598, A Guide for the Preparation of Aircraft Maintenance Documentation in the International Aerospace Maintenance Language, Issue 1, September 1995

[Androutsopoulos 95]　I. Androutsopoulos, G. D. Ritchie, P. Thanisch, Natural Language Interfaces to Databases – An Introduction, Journal of Natural Language Engineering, vol.1, no.1, Cambridge University Press, 1995

[Börger & Rosenzweig 94]　E. Börger, D. Rosenzweig, A Mathematical Definition of Full Prolog, Science of Computer Programming, 1994

[Capindale & Crawford 89]　R. A. Capindale, R. G. Crawford, Using a natural language interface with casual users, International Journal Man-Machine Studies, 32, pp. 341-362, 1989

[Covington 94 ]　M. A. Covington, GULP 3.1: An Extension of Prolog for Unification-Based Grammar, Report AI-1994-06, Artificial Intelligence Center, University of Georgia, 1994

[Covington et al. 88]　M. A. Covington, D. Nute, N. Schmitz, D. Goodman, From English to Prolog via Discourse Representation Theory, Research Report 01-0024, Artificial Intelligence Programs, University of Georgia, 1988

[Fuchs & Fromherz 94]　N. E. Fuchs, M. P. J. Fromherz, Transformational Development of Logic Programs from Executable Specifications, in C. Beckstein, U. Geske (eds.), Entwicklung, Test und Wartung deklarativer KI-Programme, GMD Studien Nr. 238, 1994

[Fuchs & Schwitter 95]　N. E. Fuchs, R. Schwitter, Specifying Logic Programs in Controlled Natural Language, CLNLP 95, Workshop on Computational Logic for Natural Language Processing, Edinburgh, 1995

[Fuchs & Schwitter 96]　N. E. Fuchs, R. Schwitter, Attempto Controlled English (ACE), CLAW 96, The first International Workshop on Controlled Language Applications, Leuven, 1996

[Ishihara et al. 92]　Y. Ishihara, H. Seki, T. Kasami, A Translation Method from Natural Language Specifications into Formal Specifications Using Contextual Dependencies, in: Proceedings of IEEE International Symposium on Requirements Engineering, 4-6 Jan. 1993, San Diego, IEEE Computer Society Press, pp. 232 - 239, 1992

| | |
|---|---|
| [Kamp & Reyle 93] | H. Kamp, U. Reyle, From Discourse to Logic, Introduction to Modeltheoretic Semantics of Natural Language, Formal Logic and Discourse Representation Theory, Kluwer Academic Publishers, Dordrecht, 1993 |
| [Kramer & Mylopoulos 92] | B. Kramer, J. Mylopoulos, Knowledge Representation, in: S. C. Shapiro (ed.), Encyclopedia of Artificial Intelligence, Wiley, 1992 |
| [Macias & Pulman 92] | B. Macias, S. Pulman, Natural Language Processing for Requirements Specifications, in: F. Redmill, T. Anderson (eds.), Safety-Critical Systems, Current Issues, Techniques and Standards, Chapman & Hall, pp. 67-89, 1993 |
| [Mitamura & Nyberg 95] | T. Mitamura, E. H. Nyberg, 3rd, Controlled English for Knowledge-Based MT: Experience with the KANT System, Center for Machine Translation, Carnegie Mellon University, Pittsburgh, 1995 |
| [Pulman 94] | S. G. Pulman, Natural Language Processing and Requirements Specification, Presentation at the Prolog Forum, Department of Computer Science, University of Zurich, February 1994 |
| [Pulman & Rayner 94] | S. Pulman, M. Rayner, Computer Processable Controlled Language, SRI International Cambridge Computer Science Research Centre, 1994 |
| [Sommerville 92] | I. Sommerville, Software Engineering, Fourth Edition, Addison-Wesley, Wokingham, 1992 |
| [Vasconcelos & Fuchs 95] | W. W. Vasconcelos, N. E. Fuchs, Opportunistic Logic Program Analysis and Optimisation – Enhanced Schema-Based Transformations for Logic Programs and their Usage in an Opportunistic Framework for Program Analysis and Optimisation, Institutsbericht 95.24, Institut für Informatik, Universität Zürich, 1995 |
| [Wojcik et al. 90] | R. H. Wojcik, J. E. Hoard, K. C. Holzhauser, The Boeing Simplified English Checker, Proc. Internatl. Conf. Human Machine Interaction and Artificial Intelligence in Aeronautics and Space, Centre d'Etude et de Recherche de Toulouse, pp. 43-57, 1990 |